\documentclass{spie}

\usepackage{cite}
\usepackage{times} 
\usepackage[utf8]{inputenc}
\usepackage[english]{babel}
\usepackage{indentfirst}
\usepackage{wrapfig}
\usepackage{graphicx}
\usepackage{subcaption}
\usepackage{varwidth}
\usepackage[ruled,linesnumbered]{algorithm2e}
\usepackage{algpseudocode}
\usepackage[colorinlistoftodos,color=pink]{todonotes}
\usepackage{xkeyval}
\usepackage{mathtext}
\usepackage{amsmath}
\DeclareMathOperator*{\argmax}{argmax}
\usepackage{gensymb}
\usepackage{todonotes}
\usepackage{multirow}
\usepackage{hyperref}
\usepackage{doi}

\SetCommentSty{mycommfont}

\graphicspath{{images/}}

\newcommand{\metaed}{0.086}
\newcommand{\mettop}{0.056}
\newcommand{\metce}{68.80}
\newcommand{\metmaxerr}{0.547\degree}

\title{A Document Skew Detection Method Using Fast Hough Transform}
\author{Bezmaternykh P.V.\supit{1,3}, Nikolaev D.P.\supit{2,3}
    \skiplinehalf
    \normalsize 
    \supit{1}Federal Research Center ``Computer Science and Control'' of Russian Academy of Sciences, Moscow, Russia;\\
    \supit{2}Institute for Information Transmission Problems of Russian Academy of Sciences, Moscow, Russia;\\
    \supit{3}Smart Engines Service LLC, Moscow, Russia;
}

\begin{document}

\maketitle

\begin{abstract}
The majority of document image analysis systems use a document skew detection algorithm to simplify all its further processing stages.
A huge amount of such algorithms based on Hough transform (HT) analysis has already been proposed.
Despite this, we managed to find only one work where the Fast Hough Transform (FHT) usage was suggested to solve the indicated problem.
Unfortunately, no study of that method was provided.
In this work, we propose and study a skew detection algorithm for the document images which relies on FHT analysis.
To measure this algorithm quality we use the dataset from the problem oriented DISEC`13 contest and its evaluation methodology.
Obtained values for $AED$, $TOP80$, and $CE$ criteria are equal to $\metaed$, $\mettop$, $\metce$ respectively.

\end{abstract}

\section{Introduction}
The Optical Character Recognition (OCR, \cite{Eikvil}) systems are widespread nowadays.
They are commonly used in digitization of archival documents, identity document recognition and voting results calculation.
One of the initial stages in these systems is the correction of a skew angle of the given document image.
The importance of this step is studied in work \cite{Bloomberg1995}.
This skew often occurs as a common result of inaccuracy during the document capturing process or other input device malfunction.
To correct this angle it firstly needs to be determined.

Not surprisingly, the indicated problem has been studied for a long time and there is plenty of well-established skew detection methods.
Among them, a group based on Hough transform analysis could be identified.
In this study, we propose and explore more computationally efficient way of document skew detection.
Instead of classic HT calculation we rely on its fast approximation, known as Fast Hough Transform \cite{Brady1998}.
The applicability of such approximation usage for the angle detection problem along with some useful insights were firstly proposed in work \cite{Nikolaev2008}, but the algorithm itself was not presented.
One of the goals of this work is to complement that study.

To measure the implementation quality of the proposed algorithm we use the public dataset from the Document Image Skew Estimation Contest (DISEC`13).
The evaluation methodology is the same as in the contest.

The rest of this work is organized as follows.
The Section \ref{sec:related_work} provides information about related work, the Section \ref{sec:proposed_algorithm} describes the proposed algorithm.
The Section \ref{sec:experiments} is dedicated to the experiments and the obtained result discussion.
Then follows the conclusion in Section \ref{sec:conclusion}.

\section{Related work}
\label{sec:related_work}
There is a number of studies on the document skew correction algorithms.
As early as 1996, the survey with annotated bibliography was published \cite{Hull1996}.
The four main groups of methods were identified therein: (i) projection profile analysis (ii) feature point distribution analysis (iii) Hough transform analysis (iv) orientation-sensitive feature analysis.
The main drawback of this study is the absence of mentioned methods accuracy comparison.
The problem of methodology for evaluating the performance of document skew estimation methods was raised around that time in the work \cite{Bagdanov1996}.
Since that, many new methods and evaluation strategies have been presented.
A notable up-to-date review of skew detection methods with their classification is presented in the work \cite{Rezaei17}.

The most basic idea behind the skew correction algorithms is the projection profile analysis.
For image $I$ of size $M \times N$ these projections are determined by the equation \eqref{eq:projection_profile}.
Usually the input image is rotated through a range of angles and these projection profiles are calculated.
This process is known as Discrete Radon Transform (DRT) calculation.
A criterion function $f$ is evaluated for every angle from the range.
This criterion represents its numerical measure and can vary a lot.
The best angle corresponds to the angle where the criterion reaches an extremum value.

\begin{equation}\label{eq:projection_profile}
\begin{split}
{\pi}_{I}^{H}(j)=\sum_{i=0}^{M}{I(i,j)},\:{\pi}_{I}^{V}(i)=\sum_{j=0}^{N}{I(i,j)}.
\end{split}
\end{equation}

The Hough transform computes a very similar kind of information.
In this work, we mainly focus our attention only on HT based methods and their accumulator space analysis.
In general, all these methods share the same approach.
Initially, the HT itself is calculated.
Then some chosen criterion $f$ is calculated for every row in its accumulator space.
The index of the row in which this criterion reaches its maximum value is selected and converted into the required angle value.
The main problem of this approach is high computational cost of the HT calculation.
To reduce this cost many techniques have been proposed.
For example, the usage of randomized HT for skew correction is described in work~\cite{Boukharouba2017}.
In the study~\cite{Yu1996} the hierarchical HT is applied for the same problem.
In fact, there is an efficient way to calculate the discrete HT approximation known as Fast Hough Transform \cite{Brady1998}
(in some works, e.g.~\cite{Singh2008}, the term ``fast'' refers to the implementation details, not the computational efficiency itself).
It has already been used in different applications of computer vision, e.g. for compensation of radial distortion~\cite{Kunina2016}, vanishing points detection~\cite{Sheshkus2018}, or for vehicle wheel tracking~\cite{Kotov2015}.
Earlier we studied some characteristics of this approximation and its applicability to the textual blocks rectification problem \cite{Limonova2017, Nikolaev2018}.
In this work, we use this approach for building more computationally efficient method for document skew detection.
The next section provides a description for such method.
\section{Proposed algorithm}
\label{sec:proposed_algorithm}

An input for the algorithm is a grayscale image $I$ (the algorithm does not require initial binarization step, but can deal with binary images as well).
Firstly, $D_h$ and $D_v$ image derivatives are calculated.
The next step is the FHT computation: $F_h$ (for ``mostly'' horizontal lines) and $F_v$ (for ``mostly'' vertical lines) \cite{Nikolaev2008}.
It is done for two directions to take into account both text lines and small vertical strokes of symbols.
Then, SSG criterion values \cite{Nikolaev2018} are computed for every row of $F_h$ and $F_v$ and stored into $C_h$ and $C_v$ vectors.
These vectors have different lengths because of the FHT calculation peculiarities.
So, they represent criterion values for different sets of angles.
To make these sets equal, $C_v$ is recalculated to the size of $C_h$.
The values of its criterion values are interpolated respectively.
After that, they can be combined into one vector $C$ by summing the appropriate values.
Finally, the $\argmax{(C)}$ is determined and is converted into the required angle.

The whole skew detection algorithm is listed in Alg. \ref{alg:main}.

\begin{algorithm}[H]
\label{alg:main}
\caption{Document skew detection algorithm using fast Hough transform}
\DontPrintSemicolon
\KwData{Input grayscale image $I$}
\KwResult{Skew angle $\alpha$}

$D_h\gets HorizontalDerivative(I)$ \\
$F_{h}\gets FastHoughTransform(D_{h})$ \\ 
$H\gets height(F_{h}), W\gets width(F_{v})$ \\
\lForEach{$i\in \mathrm{rows}(F_h)$} {\\
\Indp
$K_{h}[i]\gets \sqrt{1+i^2/(H-1)^2}$ \\
$C_{h}[i]\gets {(K_{h}[i])^{3}\cdot SSG_{FHT}(F_{h}[i])}$
}
\Indm
\BlankLine
$D_v\gets VerticalDerivative(I)$ \\ 
$F_v\gets FastHoughTransform(D_v)$ \\
\lForEach{$i\in \mathrm{rows}(F_v)$} {\\
\Indp
$K_{v}[i]\gets \sqrt{1+i^2/(W-1)^2}$ \\
$C_{v}[i]\gets {(K_{v}[i])^{3}\cdot SSG_{FHT}(F_{v}[i])}$
}
\Indm
\BlankLine
$C_{v}\gets Recalculate(C_{v})$ \\
$C\gets Combine(C_{h}, C_{v})$ \\
$I_{res}\gets \argmax{(C)}$ \\ 
$\alpha\gets \arctan{(skew(I_{res}))}$ \\
\Return $\alpha$
\end{algorithm}

\section{Experiments}
\label{sec:experiments}

\subsection{Data description and evaluation methodology}
\begin{figure}[h]
\centering
\includegraphics[width=\textwidth]{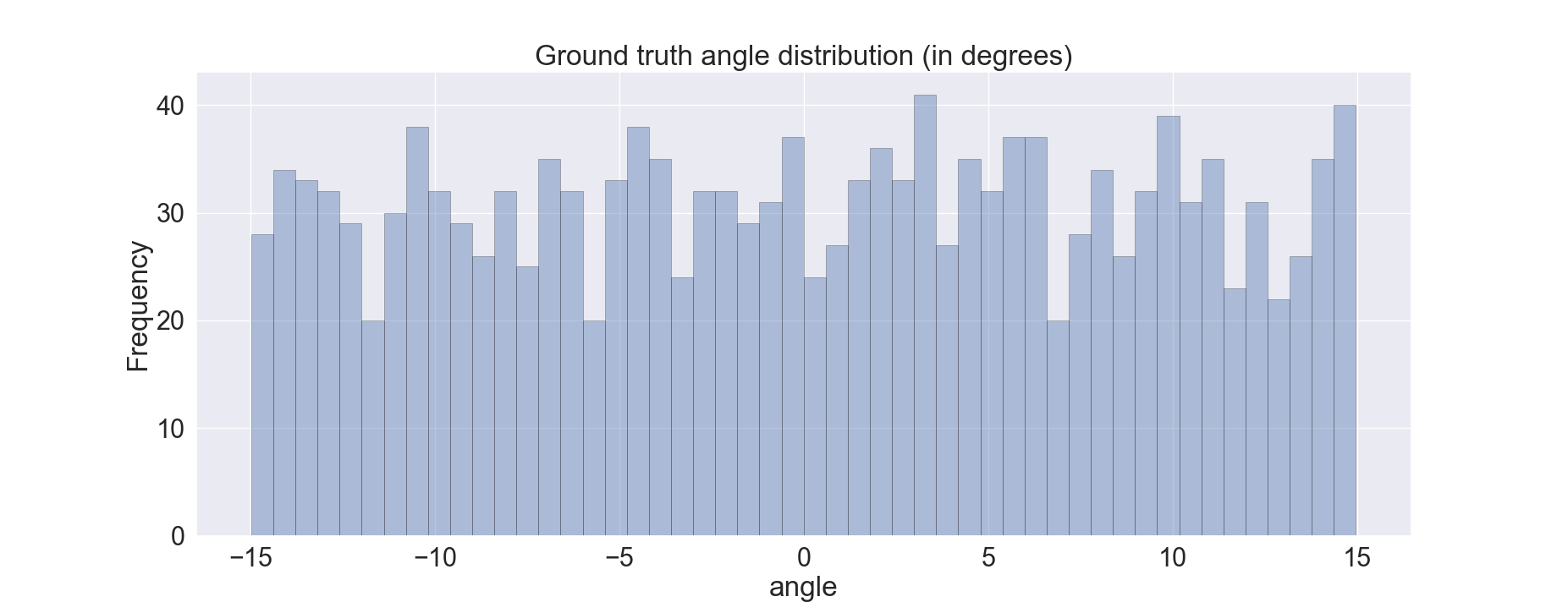} 
\caption{The histogram of angle values for the DISEC'13 dataset}
\label{fig:ground_truth}
\end{figure}

To measure the quality of the proposed algorithm we use the well-known benchmark dataset from DISEC'13 contest.
Its detailed description is provided in work \cite{DISEC}.
It consists of 155 unique images.
For each unique image 10 rotated samples are generated.
These rotation angles are randomly selected from the limited range $(-15\degree,+15\degree)$.
The ground truth angle value is established manually for every unique image once and then recalculated in accordance with mentioned randomly selected angles.
In the \figurename~\ref{fig:ground_truth} the angle values distribution for the dataset is presented.
The whole dataset is publicly available
\footnote{\url{https://www.iit.demokritos.gr/~alexpap/DISEC13/icdar2013_benchmarking_dataset.rar}}.

The DISEC'13 organizers also provided three evaluation measures: (i) average error deviation (AED), (ii) average error deviation for the top 80\% results (TOP80), (iii) correct estimation percentage (CE).
All these measures are based on absolute difference between the ground truth values and the calculated ones.
The threshold value is set to 0.1\degree.

\subsection{Results}
\label{lbl:results}

\begin{figure}[ht]
\centering
\includegraphics[width=\textwidth]{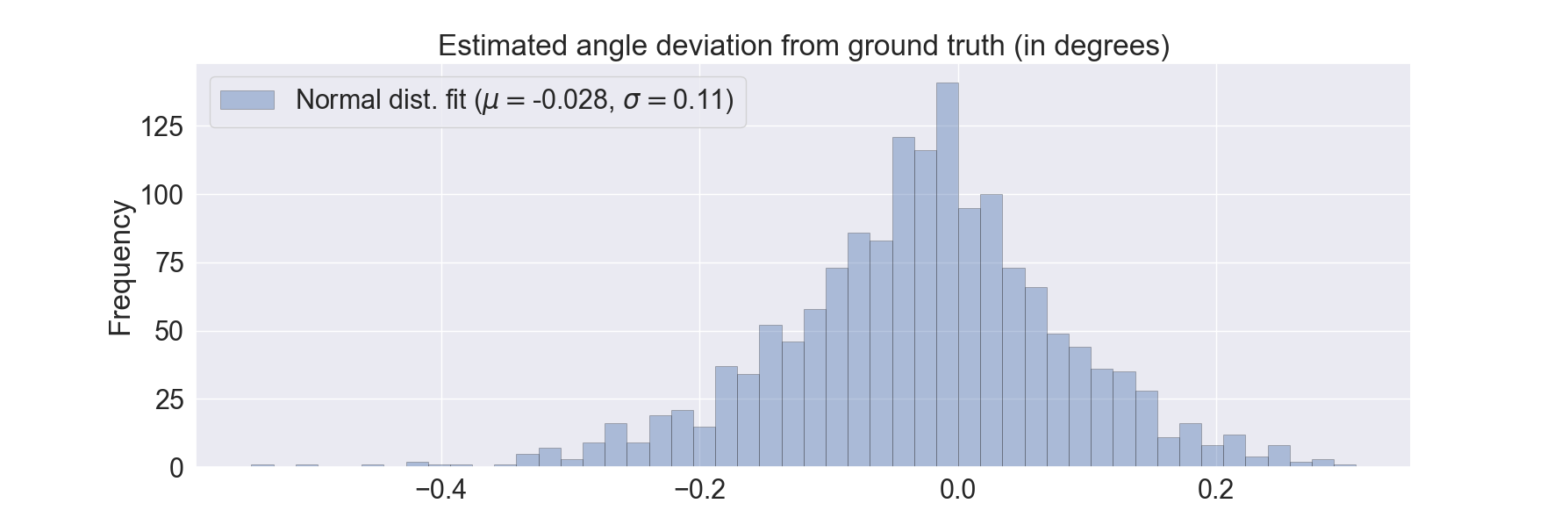} 
\caption{The histogram of errors distribution for the DISEC'13 dataset}
\label{fig:errors_distribution}
\end{figure}

In Table~\ref{tbl:result} the best methods results from the original contest are presented alongside with our implementation results.
Our method could be ranked third among the mentioned algorithms.
The top method presented in work \cite{Fabrizio2014} is based on a special preprocessing step usage followed by the Fourier analysis.
The method ranked second \cite{Koo2013} finds lines in the image in a special way and estimates the skew using the weighted votes from these extracted lines.
The LRDE-EPITA-b method is also based on a preprocessing technique with final standard Hough transform calculation for the skew angle detection.
The value of CE measure for our algorithm is significantly worse than the corresponding value for the LRDE-EPITA-a method.
The maximum angle error on the dataset is equal to~\metmaxerr, so the method does not produce serious errors.
The \figurename~\ref{fig:errors_distribution} presents the histogram of errors distribution on the benchmark dataset.

\begin{table}[ht]
    \centering
    \begin{tabular}{|c|c|c|c|c|}
        \hline
        \textbf{Method}   & \textbf{AED} (\degree) & \textbf{TOP80} (\degree) & \textbf{CE}(\%) & \textbf{DISEC rank} \\
        \hline
        \hline
        LRDE-EPITA-a & 0.072 & 0.046 & 77.48 & 1 \\ 
        Ajou-SNU     & 0.085 & 0.051 & 71.23 & 2 \\
        \textbf{Proposed} & \textbf{\metaed} & \textbf{\mettop} & \textbf{\metce} & - \\
        LRDE-EPITA-b & 0.097 & 0.053 & 68.32 & 3 \\
        Gamera       & 0.184 & 0.057 & 68.90 & 4 \\
        CVL-TUWIEN   & 0.103 & 0.058 & 65.42 & 5 \\
        \hline
    \end{tabular}
    \caption{Measurement results for the DISEC`13 dataset}
    \label{tbl:result}
\end{table}

We grouped all the samples for every unique image and measured criteria values for these groups.
We selected the results for the worst 10 groups and presented them in Table~\ref{tbl:groups}.
For these groups we also provided $MAX$, $MIN$, and $RANGE$ angle error values.

\begin{table}[ht]
\centering
\begin{tabular}{|ccccccc|}
\hline
\textbf{Index} & \textbf{AED}(\degree) & \textbf{TOP80}(\degree) & \textbf{CE}(\%) & \textbf{MAX}(\degree) & \textbf{MIN}(\degree) & \textbf{RANGE}(\degree)\\
\hline

68 &  0.307 &  0.299 &  0.0 &  0.340 &  0.266 &  0.074 \\
74 &  0.279 &  0.249 &  0.0 &  0.417 &  0.125 &  0.292 \\
18 &  0.260 &  0.235 &  0.0 &  0.406 &  0.124 &  0.282 \\
61 &  0.256 &  0.245 &  0.0 &  0.315 &  0.211 &  0.104 \\
84 &  0.253 &  0.244 &  0.0 &  0.305 &  0.202 &  0.103 \\
20 &  0.246 &  0.222 &  0.0 &  0.355 &  0.109 &  0.246 \\
40 &  0.243 &  0.232 &  0.0 &  0.289 &  0.212 &  0.077 \\
99 &  0.237 &  0.225 &  0.0 &  0.308 &  0.182 &  0.126 \\
35 &  0.229 &  0.222 &  0.0 &  0.258 &  0.188 &  0.070 \\
65 &  0.228 &  0.220 &  0.0 &  0.265 &  0.183 &  0.082 \\
\hline
\end{tabular}
\caption{Inner group measurement results for DISEC`13 dataset}
\label{tbl:groups}
\end{table}

\begin{figure}[ht]
\centering
\includegraphics[width=\textwidth,height=6cm]{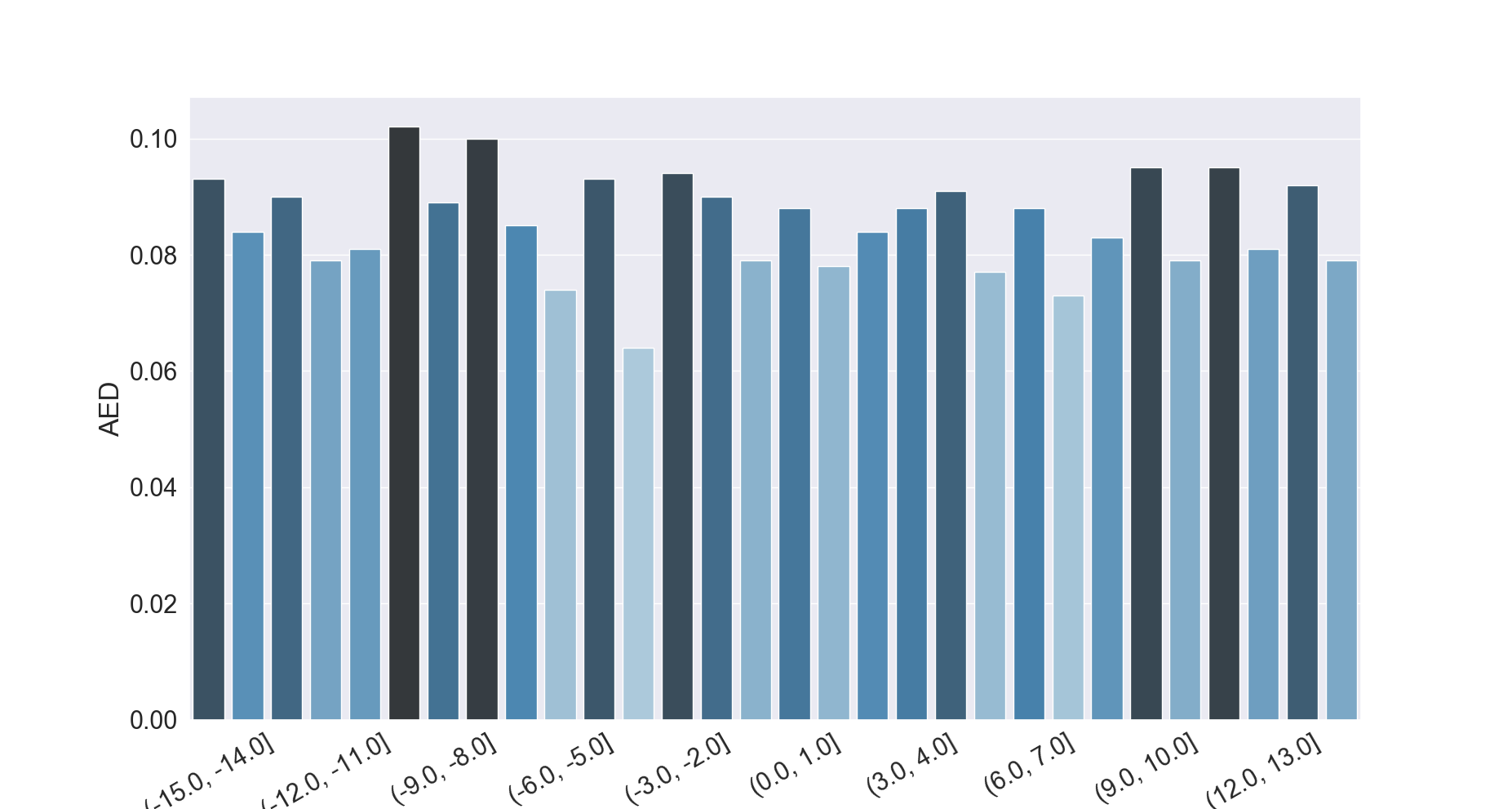} 
\caption{The AED value for every degree level for the DISEC'13 dataset}
\label{fig:grouped_by_angle}
\end{figure}

We also grouped all the samples in ranges of one degree.
We measured the AED values for these groups to confirm that proposed method error is independent of original ground truth angle value (see \figurename~\ref{fig:grouped_by_angle}).
The proposed method is straightforward and doesn't require any preprocessing step.

To understand how changing from classical image rotation and projection calculation scheme to FHT affects the running time, we measured it for both these transforms.
For image of size $1095\times894$ pixels from DISEC dataset and number of projections equal to 3975 (for this image it is an exact number of projections required for the FHT calculation for both vertical and horizontal directions) the running time for the FHT is equal to 45 $\mu$s and for the DRT it is about 21000 $\mu$s, so the performance gain is essential.
The testing machine runs under the Ubuntu 18.04 OS with
AMD Ryzen 7 1700 processor (8 cores), SSD and 16 GiB of RAM. 

\section{Conclusion}
\label{sec:conclusion}
Document image analysis is still a relevant topic in computer vision domain.
In this work, we propose an algorithm for document skew detection based on Fast Hough Transform analysis.
It does not require initial binarization step for the input image and uses FHT calculation for both ``mostly'' horizontal and vertical lines.
The reduced computational cost of FHT calculation expands the scope of its applicability.
The implementation of our algorithm managed to reach high quality on the specialized DISEC'13 dataset, but it could be enhanced further.
The maximum angle error on the dataset is equal to~\metmaxerr. 
Obtained values for criteria $AED$, $TOP80$, and $CE$ are equal to $\metaed$, $\mettop$, $\metce$ respectively.
To achieve better results we are going to use document preprocessing techniques.
We are also planning to investigate how this method works on real datasets with skewed document images (e.g. on MIDV-500 ``extra'' dataset \footnote{\url{ftp://smartengines.com/midv-500/extra/01_extra_scan.zip}}).

\acknowledgements
The reported work was partially funded by Russian Foundation for Basic Research (projects 17-29-03170 and 17-29-03370).


\bibliographystyle{spiebib}
\bibliography{main}

\end{document}